\documentclass[10pt,twocolumn,letterpaper]{article}

\usepackage{iccv}
\usepackage{times}
\usepackage{epsfig}
\usepackage{graphicx}
\usepackage{amsmath}
\usepackage{amssymb}
\usepackage{amsthm}
\usepackage{mathrsfs}
\usepackage[lined,boxed,commentsnumbered,ruled]{algorithm2e}
\usepackage{diagbox}
\usepackage{multirow}

\usepackage[pagebackref=true,breaklinks=true,letterpaper=true,colorlinks,bookmarks=false]{hyperref}

\iccvfinalcopy % *** Uncomment this line for the final submission

\def\iccvPaperID{4096} % *** Enter the ICCV Paper ID here

\ificcvfinal\fi

\begin{document}

%%%%%%%%% TITLE
\title{Prompt Tuning Inversion for Text-Driven Image Editing Using Diffusion Models}

\author{
Wenkai Dong\textsuperscript{1}\quad Song Xue\textsuperscript{1} \quad Xiaoyue Duan\textsuperscript{1, 2} \quad Shumin Han\textsuperscript{1}\\
\textsuperscript{1}Baidu VIS\\
\textsuperscript{2}Beihang University\\
{\tt\small \{dongwenkai, xuesong06, duanxiaoyue, hanshumin\}@baidu.com}
}

\maketitle
% Remove page # from the first page of camera-ready.
\ificcvfinal\thispagestyle{empty}\fi

%%%%%%%%% ABSTRACT
\begin{abstract}
Recently large-scale language-image models (e.g., text-guided diffusion models)
% , like text-guided diffusion models, 
have considerably improved the image generation capabilities
% , and are able 
to generate photorealistic images in various domains. Based on this success, current image editing methods use texts to achieve intuitive and versatile modification of images. To edit a real image using diffusion models, one must first invert the image to a noisy latent from which an edited image is sampled with a target text prompt. 
However, most methods lack one of the following: user-friendliness (e.g., additional masks or precise descriptions of the input image are required), generalization to larger domains, or high fidelity to the input image.
% However, most methods are limited to one of the following: user-friendliness (e.g., additional masks or accurate descriptions of the input image are required), specific domains or high fidelity. 
In this paper, we design an accurate and quick inversion technique, Prompt Tuning Inversion,
for text-driven image editing.
% for the text-driven editing of the image. 
Specifically, our proposed editing method consists of a reconstruction stage and an editing stage. In the first stage, we encode the information of the input image into a learnable conditional embedding via Prompt Tuning Inversion. In the second stage, we apply classifier-free guidance to sample the edited image, where the conditional embedding is calculated by linearly interpolating between the target embedding and the optimized one obtained in the first stage. This technique ensures a superior trade-off between editability and high fidelity to the input image of our method. For example,
% With this novel inversion technique, our method is user-friendly and achieves a superior trade-off between editability and fidelity. For example, 
we can change the color of a specific object while preserving its original shape and background under the guidance of only a target text prompt. 
% Our proposed image editing method consists of a reconstruction stage and an editing stage. 
% In the first stage, we encode the information of the input image into a learnable conditional embedding via Prompt Tuning Inversion.}
% % In the first stage, we introduce a learnable conditional embedding and  encode the information of the input image via Prompt Tuning Inversion. 
% In the second stage, we apply classifier-free guidance to sample the edited image, where the conditional embedding is calculated by linearly interpolating between the target embedding and the optimized embedding obtained in the first stage}. 
Extensive experiments on ImageNet demonstrate the superior editing performance of our method compared to the state-of-the-art baselines.
\end{abstract}

%%%%%%%%% BODY TEXT
\section{Introduction}
Text-based image editing, a long-standing problem in image processing, aims to modify an input image to align its visual content with the target text prompts. It has drawn increasing attention in recent years and many methods built upon text-to-image generation have been developed. In past years, GAN-based image editing methods~\cite{park2019semantic,patashnik2021styleclip,vinker2021image,wang2018high} achieve impressive results due to the powerful generation abilities of GANs~\cite{reed2016generative, liang2020cpgan, qiao2019mirrorgan, zhu2019dm}. However, these methods only work well in domains where the models are trained.
% the domain where their models are trained. 
More recently, diffusion models such as DDPM~\cite{ho2020denoising} and score-based generative models~\cite{song2020score} have demonstrated competitive or even 
better capability
% higher quality 
of generating images 
compared
% compare 
to 
VAE-, GAN-, flow- and autoregressive-based
% VAE, GAN, flow, and autoregressive-based 
% generation 
models~\cite{razavi2019generating,goodfellow2020generative,rezende2015variational,esser2021taming}. Especially, large-scale language-image models (LLIMs), such as Imagen~\cite{saharia2022photorealistic}, DALL-E2~\cite{ramesh2022hierarchical} and Stable Diffusion~\cite{rombach2022high}, have attracted unprecedented attention from the research community and public society. With the help of large-scale pre-trained language models~\cite{radford2021learning,devlin2018bert}, LLIMs can generate high-fidelity images well aligned with the provided text prompts without further fine-tuning. To fully leverage the 
generation and generalization capabilities
% generative capacities and generalities 
of LLIMs, we aim to develop a text-driven image editing method based on open-sourced LLIMs, \emph{e.g.}, Stable Diffusion~\cite{rombach2022high}.

\begin{figure}[t]
    \centering
    \includegraphics[width=0.9\linewidth]{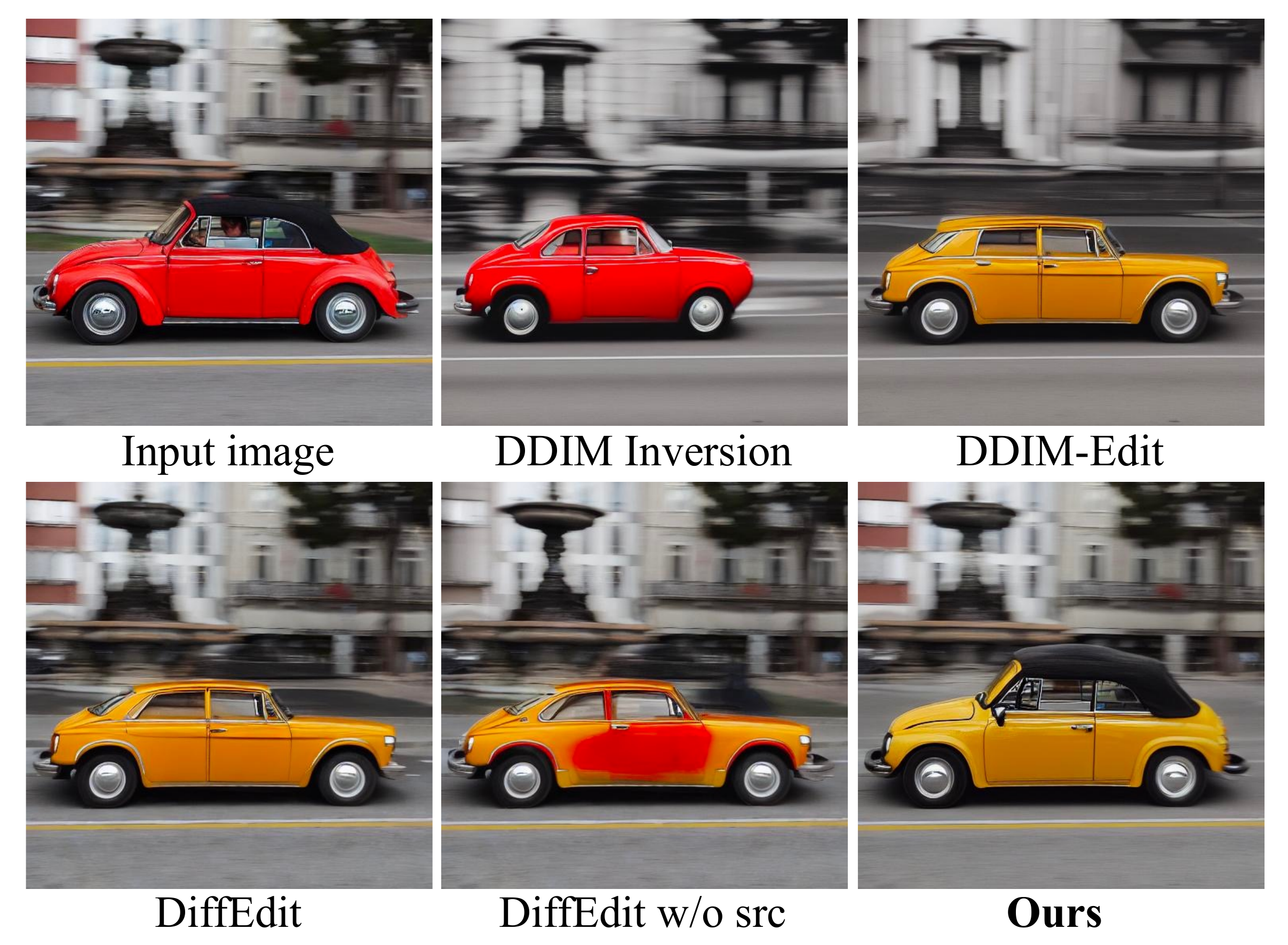}
    \caption{
    Illustration of different methods in editing the color of the car. Methods based on the original DDIM inversion (\emph{i.e.}, DDIM-Edit, DiffEdit and DiffEdit w/o src) cannot preserve the shape of the car. In contrast, our method successfully changes the color while preserving the structural information. The target text is ``a yellow car''. The source text is ``a red car'' for DiffEdit.}
    % An example of changing the color of the car. Methods with initial DDIM inversion (DiffEdit, DDIM-Edit, DiffEdit w/o src) cannot preserve the shape of the car. In contrast, ours works well.
    \label{F1_1}
\end{figure}

\textbf{Editability} and \textbf{fidelity} are two essential requirements of image editing tasks.
% Image editing has two needs, editability and high fidelity to the input image. 
The former requires that the edited images are supposed to contain visual contents well aligned with the corresponding textual contents provided in the target prompts,
% The former means that modified images are supposed to contain visual contents align with corresponding textual content in target prompts.  
while the latter expects that areas other than the edited parts should stay as close to those of the input image as possible.
% The latter means that the parts of the modified image other than the editing area should be consistent with the input image. 
For example, when modifying the color of a specific object, its other attributes (\emph{e.g.}, size and shape) are expected to be preserved.
% Sometimes modifying the color of a specific object needs to preserve the original shape.
As shown in Fig.~\ref{F1_1}, given an image of a red car and the target text prompt (``a yellow car''), the desired edited image should contain a yellow car while keeping the background as well as the car's size and shape unchanged.
% For instance, as shown in Figure, given an image of a red car and the query a yellow car, a desired modified image should contain a yellow car while keeping the background and the car's shape as similar as possible to the input image.  
To achieve this editing,
% For editing a real image, 
the simplest way is to first invert the image to a noisy latent via the reversed deterministic DDIM sampling process~\cite{song2020denoising}, and then obtain the edited image via the deterministic DDIM sampling process with the guidance of the target prompt embedding. We refer to this approach as ``DDIM-Edit'' in 
% Fig.~\ref{F1_1} and the rest of 
our paper.
% the simplest way, referred to as DDIM edit in this paper, is to first invert the image to a noisy latent via the reversed deterministic DDIM sampling process. Then we can obtain the edited image by using the target embedding to guide the deterministic DDIM sampling.
Although this approach successfully turns the color of the car to yellow (see ``DDIM-Edit'' in Fig.~\ref{F1_1}), the background and the shape of the car change drastically,
% However, as shown in Figure, although the color is changed to yellow, the background and the shape of the car change significantly, 
which obviously fails to meet the requirement of high fidelity.  The reason lies in that the deterministic DDIM sampling process cannot be reversed perfectly in practice.
A slight error is amplified by a large classifier-free guidance scale, and is accumulated in each sampling step, which consequently results in a significantly different image. 

To improve fidelity, some methods 
consider the image editing tasks as inpainting tasks, which
% convert the image editing task to the image inpainting task and 
require users to explicitly provide masks of the inpainting regions~\cite{avrahami2022blended,lugmayr2022repaint}. With the mask prior,
the background can remain the same, but masking out image contents also removes important structural information that is helpful in the editing process,
% although the background remains the same, 
% masking the image content removes important structural information, which should be used in image editing, 
leading to unsatisfactory editing results.
% leading to a completely different edited object. 
Moreover, asking 
users
% the user 
to 
provide masks
% mask images 
is cumbersome and not suitable for quick and intuitive text-driven image editing.
As a solution,
DiffEdit~\cite{couairon2022diffedit} presents an algorithm that can automatically generate a mask
given a target text prompt to locate the region to be edited.
% where the region should be edited given a target text. 
However, the editability of DiffEdit largely depends on 
DDIM-Edit,
% DDIM edit, 
which may fail to preserve the structural information of the edited object, \emph{e.g.}, 
the shape of the car (see ``DiffEdit'' in Fig.~\ref{F1_1}).
% the car's shape in Figure changes. 
Moreover, to generate an accurate mask,  DiffEdit requires a precise text description of the input image (referred to as ``source text''), 
hampering the editing efficiency.
% hampering the quick editing process. 
Without the source text (see ``DiffEdit w/o src'' in Fig.~\ref{F1_1}),
% As shown in Figure, without the reference text, 
the automatically generated mask 
cannot locate the body of the car accurately,
% is not accurate, 
further decreasing editability. 

In this work, we aim to propose an image editing method to mitigate all the above problems,
\emph{i.e.}, the method should be user-friendly, generalizable to various domains, and generate edited images with high fidelity.
% The method owns the following features: user-friendly, generability, and high-fidelity.  
Specifically, for a quick and intuitive text-based method, users only need to provide an input image and the corresponding target text prompts, 
without the need for a mask or a source text describing the input image.
% neither a mask nor reference text describing the input image. 
Secondly, the method
should be able to
% can 
operate on real images from various domains. 
Thirdly, the objects should be precisely edited with the background preserved. In some cases, only certain attributes of the objects should be modified, while other attributes are supposed to be left untouched.
% Thirdly, only specific objects should be modified and the background is preserved. In some cases, the structure of the object should not change significantly. 

To achieve these merits, we believe that image editing needs a new inversion method based on diffusion models to reconstruct the input image. Inspired by the classifier-free guidance~\cite{ho2022classifier} and textual-inversion methods~\cite{mokady2022null}, we propose a \textbf{Prompt Tuning Inversion} method to encode the information of the input image into a conditional embedding. More specifically, we first apply DDIM inversion to the input image
latent
to obtain a sequence of noisy ones. These noisy latents can be taken as a 
prior
trajectory for 
reconstructing the original image.
% reconstruction. 
Then, we introduce a learnable embedding in the sampling process.
% In the sampling process, we introduce a learnable embedding.
The diffusion model reconstructs the input image step by step along the trajectory conditioned on this embedding while optimizing it at the same time. In this way, the contents of the input image are learned in the embedding.
% on which the diffusion model is conditioned and reconstructs the input image step by step along the trajectory while optimizing the embedding. 
Finally, we obtain a new conditional embedding by linearly interpolating between the optimized embedding and the target embedding, resulting in a representation that combines both the structural information of the input image and the visual content of the target text. 

Overall,
% From an overall view, 
our proposed
% image editing 
method consists of two stages. In the first stage, we encode the information of the input image into a learnable conditional embedding via prompt tuning in the reconstruction process.
In the second stage, a new conditional embedding is computed by linearly interpolating between the target embedding and the optimized one obtained in the first stage, which boosts a trade-off between editability and fidelity. The classifier-free guidance is then applied to sample the edited image.
% In the second stage, we calculate a new conditional embedding by linearly interpolating between the target embedding and the optimized one. Then we apply classifier-free guidance to sample the edited image.
In sum, our contributions are as follows:
\begin{itemize}
    % \vspace{-1mm}
    \item We propose a user-friendly text-driven image editing method
    which requires only an input image and a target text for editing, without any need for user-provided masks or source descriptions of the input images.
    % which requires only an input image and a target text for editing, and}
    % % which 
    % achieves a 
    % satisfactory trade-off}
    % % good balance 
    % between editability and fidelity. 
    % Our method requires users to provide only input images and target texts.
    % \vspace{-1mm}
    \item We propose a Prompt Tuning Inversion method for diffusion models which can quickly 
    and accurately
    reconstruct the original image,
    providing a strong basis for sampling edited images with high fidelity to the inputs.
    % \vspace{-1mm}
    \item We compare against the state-of-the-art methods both qualitatively and quantitatively, and show that our method outperforms these works in terms of the trade-off between editability and fidelity.
\end{itemize}

%-------------------------------------------------------------------------
% \vspace{-1mm}
\section{Related work}

\noindent\textbf{Text-to-image synthesis.} Text-guided synthesis has been widely adopted for image generation~\cite{esser2021taming, wang2022clip, chang2022maskgit, rombach2022high}.
% Text guided synthesis methods has been widely used as a new image generation method. 
% Representative methods include}
% % Representative articles mainly include 
% VQ-GAN~\cite{esser2021taming}, CLIP-GEN~\cite{wang2022clip}, MaskGIT~\cite{chang2022maskgit}, and Stable Diffusion~\cite{rombach2022high}, \emph{etc}. 
% By feeding the corresponding texts, these methods are capable of generating almost any image.}
% they can generate almost any image, just feed the corresponding text. 
Works based on generative adversarial networks (GANs)~\cite{reed2016generative, liang2020cpgan, qiao2019mirrorgan, zhu2019dm} have been proposed for text-to-image synthesis.
% GAN-INT-CLS~\cite{reed2016generative} is the first to use a conditional GAN~\cite{goodfellow2020generative}
% % formulation 
% for text-to-image generation, after which~\cite{liang2020cpgan,qiao2019mirrorgan,zhu2019dm} are proposed to further improve the generation quality. 
% However, the training of GANs is unstable. VQGAN~\cite{esser2021taming} greatly improves the performance of image generation by training an encoder to compress images into a low-dimensional discrete latent space and fitting the hidden variables. 
CLIP-based methods~\cite{crowson2022vqgan, wang2022clip} have also been proposed to utilize the language-image priors from a pre-trained CLIP~\cite{radford2021learning} model to generate images from texts.
% VQGAN-CLIP~\cite{crowson2022vqgan} further adopts CLIP-guided VQGAN~\cite{esser2021taming} to generate various styles of images from text prompts. However, these methods require corresponding image-text pairs. CLIP-GEN~\cite{wang2022clip} proposes a self-supervised scheme using language-image priors extracted from a pre-trained CLIP model. 
% MaskGIT~\cite{chang2022maskgit} proposes a new paradigm using a bidirectional transformer decoder, further speeding up the auto-regressive decoding process.
Recently, works~\cite{dhariwal2021diffusion,ho2020denoising,ho2022cascaded, nichol2021improved} based on the Diffusion Probabilistic Models (DPM)~\cite{sohl2015deep} have achieved state-of-the-art results in text-to-image synthesis. 
% However, most works consider continuous diffusion models over raw image pixels, which are computationally demanding. 
% VQ-Diffusion~\cite{nichol2021improved} proposes a discrete diffusion model, which adopts VQVAE~\cite{van2017neural,razavi2019generating} to encode images into discrete tokens and performs the diffusion process in a discrete space.
% By estimating the probability of each discrete token, it can obtain high-quality images.
% The discrete diffusion model is described in VQ-Diffusion~\cite{nichol2021improved}. VQ-Diffusion utilizes VQVAE~\cite{van2017neural,razavi2019generating} to encode images into discrete tokens, then performs a diffusion process in a discrete space, and can estimate the probability of each discrete token, so it can obtain high-quality images. 
Among these works, Latent Diffusion Model (LDM)~\cite{rombach2022high} trains DPM in the latent space using a powerful pre-trained auto-encoder, and introduces a cross-attention layer into the model architecture, thus turning the diffusion model into
% Besides that, LDM~\cite{rombach2022high} applies DM training to the latent space of a powerful pre-trained autoencoder and introduces a cross-attention layer into the model architecture, thus turning the diffusion model into 
a powerful and flexible generator with greatly improved visual fidelity. 
Our work of image editing is based on LDM~\cite{rombach2022high} thanks to its powerful image generation capability.
% Due to the powerful image generation ability of the diffusion model, our image editing work will be based on LDM~\cite{rombach2022high}.

\noindent\textbf{Image editing.} Image editing with generative adversarial networks (GANs) has been studied extensively~\cite{park2019semantic,patashnik2021styleclip,vinker2021image,wang2018high}.
% Many of these works utilize the latent space of pre-trained GANs to conduct image manipulations from the modified latent representations~\cite{harkonen2020ganspace,lang2021explaining,shen2020interpreting}. 
% Typical techniques have been proposed for applying such manipulations, \emph{e.g.}, designing better encoders~\cite{richardson2021encoding,tov2021designing}, discovering better latent directions~\cite{goetschalckx2019ganalyze,jahanian2019steerability,shen2020interpreting}, fine-tuning the parameters of the model~\cite{pan2021exploiting,roich2022pivotal} and so on. 
Some other techniques also leverage the image-text alignment capability of CLIP~\cite{radford2021learning} and transfer it to the framework of GANs~\cite{abdal2022clip2stylegan,stap2020conditional,xia2021tedigan}.
% However, these methods are computationally intensive, and are often limited to specific domains, making it difficult for them to generalize to larger and more diverse datasets~\cite{karras2020training}. 
More recently, the development of diffusion models~\cite{ho2020denoising,sohl2015deep,song2019generative} provides a more flexible design space than GANs for the editing task, while following a simpler training setup (\emph{e.g.}, SDEdit~\cite{meng2021sdedit} and ILVR~\cite{choi2021ilvr}).
% SDEdit~\cite{meng2021sdedit} performs image editing by reversing the input image perturbed by target noise or other corruptions using the diffusion model, but loses partial fidelity to the original image. ILVR~\cite{choi2021ilvr} guides the denoising process with the constraint that the decoded image should stay close to the base image. 
Textual Inversion~\cite{gal2022image} and Dream-Booth~\cite{ruiz2022dreambooth} demonstrate the capability to generate diverse images with unique object characteristics by fine-tuning the diffusion model with multiple images. 
% DiffusionCLIP~\cite{kim2022diffusionclip} provides gradients for image manipulation leveraging CLIP~\cite{radford2021learning}. 
Imagic~\cite{kawar2022imagic} and UniTune~\cite{valevski2022unitune}, which are based on the powerful Imagen model~\cite{saharia2022photorealistic}, also show impressive editing performance. However, the above methods require restrictive fine-tuning of the pre-trained model, and thus may not fully leverage the generalization ability of the pre-trained model due to overfitting or language drift. Other methods~\cite{avrahami2022blended,nichol2021glide} require a user-provided mask to guide the diffusion process, making it hard for them to be interactive. To achieve text-only interactive editing, some optimization-free methods have been proposed recently (\emph{e.g.}, Prompt-to-Prompt~\cite{hertz2022prompt} and DiffEdit~\cite{couairon2022diffedit}) to automatically infers a mask before editing. 

\noindent\textbf{Inversion.} In the GAN literature, the inversion process requires one to find a corresponding latent representation of the given image~\cite{zhu2016generative, xia2022gan}. This process has been extensively studied for GANs~\cite{abdal2019image2stylegan, zhu2020improved, gu2020image,richardson2021encoding,pidhorskyi2020adversarial,tov2021designing}.
% including optimization-based approaches~\cite{abdal2019image2stylegan, zhu2020improved, gu2020image} or encoder-based techniques~\cite{richardson2021encoding, pidhorskyi2020adversarial, tov2021designing}. 
For diffusion models, the inversion requires to find a noise map and a conditional vector corresponding to a generated image but simply adding noise and denoising it may arouse the problem that the image content can be changed drastically. Works~\cite{choi2021ilvr, dhariwal2021diffusion, ramesh2022hierarchical} have been proposed to improve the inversion process. However, it is still challenging for these methods to generate new instances of a given example while maintaining fidelity. Textual Inversion~\cite{gal2022image} and Dream-Booth~\cite{ruiz2022dreambooth} propose to learn concepts from images through textual inversion by either directly optimizing the embedding of the textual concept or fine-tuning the diffusion models, which can be computationally inefficient. Null-Text Inversion~\cite{mokady2022null} modifies the unconditional textual embedding that is used for classifier-free guidance instead of the input text embedding, which enables applying prompt-based editing without the cumbersome tuning of the model parameters.
Different from these methods, our method encodes the information of the input image into a learnable conditional embedding, which provides a helpful prior in sampling edited image with high fidelity to the input image.

%------------------------------------------------------------------------
\section{Methodology}
%------------------------------------------------------------------------

\begin{figure*}[!htp]
    \centering
    \includegraphics[width=0.75\textwidth]{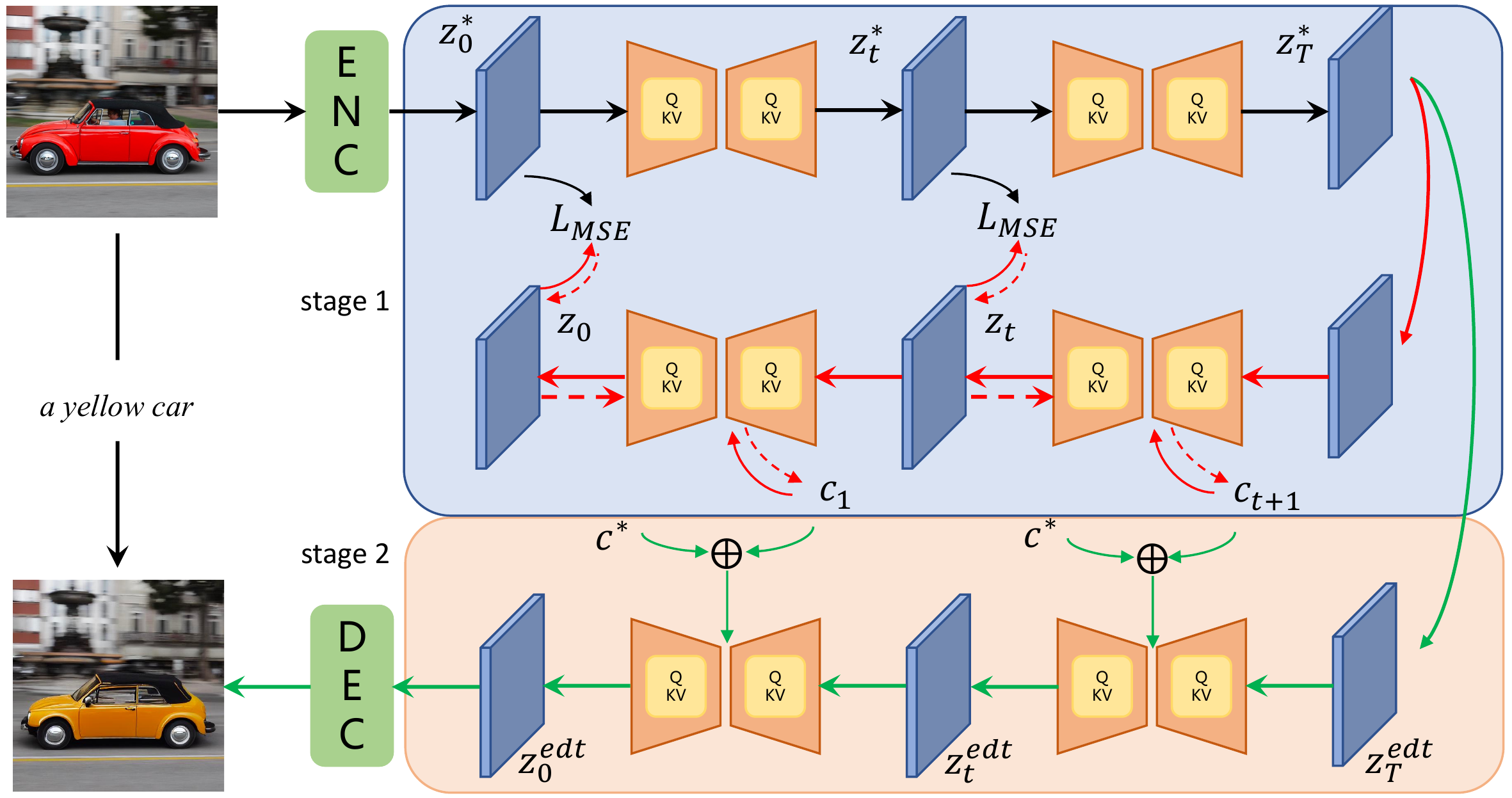}
    \caption{An overview of our proposed image editing method. \textbf{Stage 1}: we first apply
    % a
    DDIM inversion
    % method 
    to the input image embedding to obtain a diffusion trajectory $\{z_t^*\}_{t=0}^T$. Then we reconstruct the input image along with the reversed trajectory by optimizing the learnable conditional embedding $c_t$. \textbf{Stage 2}: we perform classifier-free guidance sampling conditioned on a linear 
    interpolation
    % combination 
    between target embedding $c^*$ and $c_t$ at each diffusion step. 
    $\bigoplus$ denotes element-wise weighted addition. Red dashes indicate the gradient flow in PTI.}
    \label{F3_2}
\end{figure*}

Given a real or synthesized image $\mathcal{I}$, we aim to edit $\mathcal{I}$ to get an edited image  $\mathcal{I^*}$ with the guidance of text. Different from existing methods which require source prompts provided by users or produced by an off-the-shelf image captioning model, our proposed editing process is guided by only target or edited prompt $\mathcal{P^*}$. 
An overview of our method is provided in Fig.~\ref{F3_2}, which
% The
% % Our 
% proposed image editing method 
consists of two stages.
In the first stage, we encode the information of the input image into a learnable conditional embedding via prompt tuning in the reconstruction process. A new conditional embedding is then computed in the second stage by linearly interpolating between the target embedding and the optimized one from the first stage, thus achieving effective editing while maintaining high fidelity. Conditioned on this interpolated embedding, the classifier-free guidance is adopted to sample the final edited image.
% \textcolor{red}{In the first stage, we encode the information of the input image into a learnable conditional embedding via prompt tuning in the reconstruction process. In the second stage, we calculate a new conditional embedding by linearly interpolating between the target embedding and the optimized one.}

% In this section, we first}
% % Next, we 
% provide a brief background of our method
% in Sec.~\ref{sec:3.1}.}
% % including latent diffusion and classifier-free guidance. 
% Then we 
% discuss}
% % present 
% the problems of existing inversion methods 
% in Sec.~\ref{sec:3.2},}
% and 
% elaborate on}
% % a detailed description of 
% our approach in 
% Secs.~\ref{sec:3.3} and~\ref{sec:3.4}. A general overview of our method is provided in Fig.~\ref{F3_2}.}
% % {\color{red}Sec. 3.2, Sec. 3.3 and Sec. 3.4. A general overview is provided in Fig. }

%------------------------------------------------------------------------
\subsection{Background and preliminaries}
\label{sec:3.1}

\noindent\textbf{Diffusion models.} Diffusion probabilistic models are designed to learn a data distribution by gradually denoising normally distributed noise, which corresponds to learning to reverse a fixed forward diffusion process:
\begin{equation}
    \label{E1}
    q(x_t|x_{t-1}) := \mathcal{N}(x_t; \sqrt{\alpha_t}x_{t-1}, (1-\alpha_{t})\textbf{I}).
\end{equation}

In the forward process, normally distributed noise is gradually added to the sample $x_{t-1}$ to obtain a more noisy variant $x_t$. The noise is dependent on a variance schedule $\alpha_t$ where $t \in {1, ..., T}$, with $T$ being the total number of steps,
$x_0$ the original image, and $x_T$ approximately the standard Gaussian noise.
The reverse process is defined with parameters $\theta$:
\begin{equation}
    \label{E2}
    p_{\theta}(x_{t-1}|x_t) := \mathcal{N}(x_{t-1}; \mu_{\theta}(x_t,t), \Sigma_{\theta}).
\end{equation}
Using a fixed variance 
$\Sigma_{\theta}$,
% $\Sigma_{\theta}(x_t, t)$, 
only the
mean value $\mu_{\theta}(x_t,t)$ needs
% means need 
to be learned. With the parameterization trick, the network $\epsilon_{\theta}$ is trained to predict
the
% artificial 
noise $\epsilon$, resulting in a loss,  where
$c$ represents the conditional embedding:
\begin{equation}
    \label{E3}
    \mathbb{E}_{t, x_0, \epsilon}[||\epsilon - \epsilon_{\theta}(x_t, t, c)||^2].
\end{equation}

In this work, we employ the deterministic DDIM sampling~\cite{song2020denoising}:
\begin{equation}
    \label{E4}
    x_{t-1} = \sqrt{\alpha_{t-1}}f_\theta(x_t, t) + \sqrt{1 - \alpha_{t-1}}\epsilon_\theta(x_t, t),
\end{equation}
where $f_\theta$ is the prediction of $x_0$ given $x_t$ at step $t$. 
Given a noisy image $x_T$, the noise is gradually removed to generate an image $x_0$ by applying Eq.~\ref{E4} for $T$ steps.
% To generate an image, given a noise $x_T$, the noise is gradually removed by using Eq.~\ref{E4} for $T$ steps. 

\noindent\textbf{Latent diffusion.} Instead of operating in the image
pixel
space, Latent Diffusion Models~\cite{rombach2022high} (LDMs)
utilize an autoencoder to learn a latent space which is perceptually equivalent to the pixel space. First, an encoder $ENC$ is
adopted
% trained 
to map a given image $x_0$ into a latent embedding $z_0$. Then a decoder $DEC$ is designed to reconstruct the input image given $z_0$, \emph{i.e.}, $DEC(ENC(x_0))\!\approx\!x_0$. The encoder downsamples the original images by a factor of
4
% four
or
8.
% eight. 
In this way, the diffusion model operates on a much smaller representation with lower
time complexity and memory burden.
% time and space complexity. 
Thus, 
for our method, we apply
% we apply our method over 
one of the state-of-the-art LDMs, 
Stable Diffusion~\cite{rombach2022high}. In the forward and reverse process described above, we only need to replace the image $x_t$
with
its latent embedding $z_t$ at each step.

\noindent\textbf{Classifier-free guidance.} Our editing method is built upon text-guided diffusion models. In Stable Diffusion, the text
$\mathcal{P}$
% $P$
is
fed into
% produced by 
a pre-trained CLIP~\cite{radford2021learning} text encoder $\tau_\theta$ to obtain its corresponding embedding and the underlying UNet model is augmented with the cross attention mechanism, which is effective for generating visual contents conditioned on the text $\mathcal{P}$. One of the key challenges in this kind of generation models is the amplification of the effect induced by the conditional text. To this end, the classifier-free guidance technique is proposed, where the prediction
for each step
is a combination of conditional and unconditional predictions. Formally, let $c=\tau_\theta(\mathcal{P})$ be the conditional embedding vector and $\varnothing=\tau_\theta(``")$ be the unconditional one, the classifier-free guidance prediction is calculated by:
\begin{equation}
    \label{E5}
    \tilde{\epsilon}_t=\epsilon_\theta(z_t, t, \varnothing) + \omega\cdot(\epsilon_\theta(z_t, t, c)-\epsilon_\theta(z_t, t, \varnothing)),
\end{equation}
where $\omega$ is the guidance scale parameter.

\subsection{Problems of DDIM inversion}
\label{sec:3.2}
Given an input image $\mathcal{I}$ and a target prompt $\mathcal{P}^*$, we aim to edit $\mathcal{I}$ to make its visual content consistent with textual content in $\mathcal{P}^*$, while preserving a maximal amount of details from $\mathcal{I}$.
% (\emph{e.g.}, small details in the background and the identity of the object within the image.) 
The above two aspects are referred to as \textbf{editability} and \textbf{input image fidelity}, respectively. 

To achieve effective editing while maintaining high fidelity, we first need to inverse the input image into an appropriate noise map, based on which the edited image can be sampled. Eqs.~\ref{E1} and~\ref{E2} show a naive way to add noise to the input image and then denoise it through the diffusion network, respectively.
% To achieve a satisfactory trade-off between editability and fidelity, 
% a noise map of the input image and an appropriate conditional embedding are needed }
% \textcolor{red}{we need to find a noise map and conditioning embedding to generate the input image, which is referred to as the \textbf{inversion} of the diffusion model.} 
% A naive way is to add noise to the input image as the forward process in Eq.~\ref{E1}, and then denoise it through the diffusion network as the reverse process in Eq.~\ref{E2}. 
However, as the sampling process is stochastic, the samples generated from the same latent can be different every time. Even if the sampling process becomes deterministic, the random noise in the forward process still makes the generated image content change significantly. To address this issue, DiffusionCLIP~\cite{kim2022diffusionclip} reverses the deterministic DDIM sampling process in Eq.~\ref{E4} based on the assumption that the ordinary differential equation (ODE) process can be reversed within the limit of small steps:
\begin{equation}
    \label{E6}
    z_{t+1} = \sqrt{\alpha_{t+1}}f_\theta(z_t, t) + \sqrt{1 - \alpha_{t+1}}\epsilon_\theta(z_t, t),
\end{equation}
where $z_t$ is the latent embedding of $x_t$.

%------------------------------------------------------------------------
\begin{figure}[t]
    \centering
    \includegraphics[width=\linewidth]{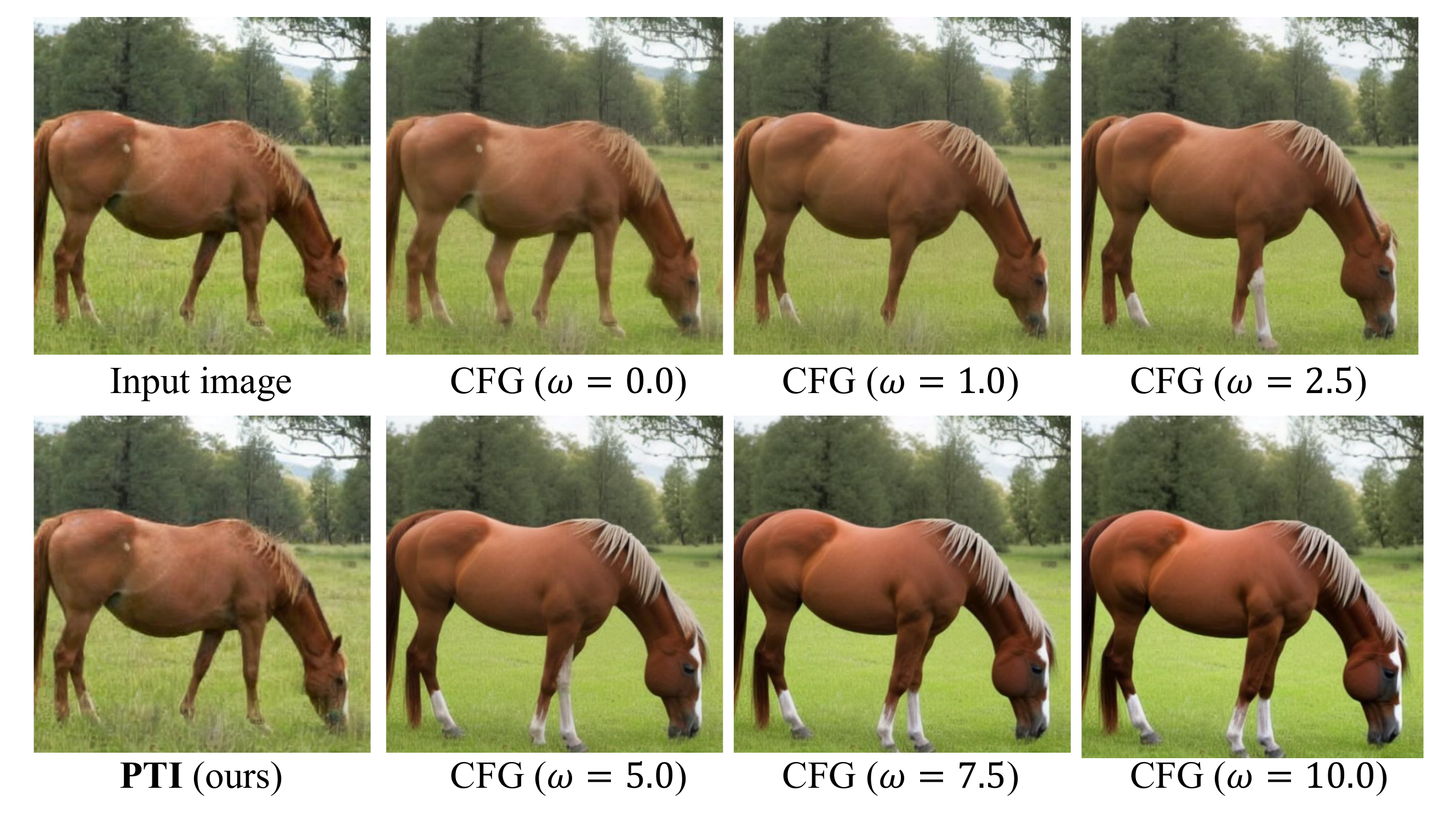}
    \vspace{-3mm}
    \caption{
    Reconstruction quality of our Prompt Tuning Inversion, and DDIM inversion with different classifier-free guidance (CFG) scales $\omega$ in the sampling process. $\omega\!=\!0$ in the forward process for all methods.}
    % Prompt Tuning Inversion (PTI), and DDIM inversion with different classifier-free guidance scales (CFG).
    \label{F3_1}
\end{figure}

\begin{table}[t]
    \centering
    \resizebox{0.88\linewidth}{!}{
    \begin{tabular}{l|c|c|c|c|c}
    \hline
    \diagbox{$\omega_{\text{enc}}$}{$\omega_{\text{dec}}$} & 0.0   & 1.0   & 2.5   & 5.0   & 7.5   \\
    \hline 
    0.0                    & \textbf{21.36} & 19.79 & 17.04 & 14.88 & 13.64 \\
    \hline 
    1.0                    & 17.16 & \textbf{21.94} & 18.22 & 15.47 & 14.02 \\
    \hline 
    2.5                    & 14.51 & 15.73 & \textbf{18.13} & 15.74 & 14.24 \\
    \hline 
    5.0                    & 11.30 & 11.48 & 11.70 & \textbf{12.12} & 12.06 \\
    \hline
    \end{tabular}
    }
    \vspace{+1mm}
    \caption{Reconstruction quality by measuring the PSNR score of DDIM inversion with different classifier-guidance scales $\omega$.
    $\omega_{\text{enc}}$ and $\omega_{\text{dec}}$ denote the guidance scale used in the DDIM forward and sampling processes, respectively.}
    % ``enc-s'' and ``dec-s'' denote the guidance scale $\omega$ used in the DDIM forward and sampling process, respectively.
    \label{T1}
\end{table}

To investigate the reconstruction performance of DDIM,
% Based on DDIM inversion, 
we first invert the latent embedding of the input image into noise maps via Eq.~\ref{E6}, and then use the deterministic DDIM sampling process in Eq.~\ref{E4} to reconstruct the input. Note that both processes are performed with unconditional diffusion models, \emph{i.e.}, the classifier-free guidance scale in Eq.~\ref{E5} is set to $\omega\!=\!0$ for both forward and reverse sampling processes.
% $\omega\!=\!0$.  
Although a slight error is incorporated in every step as ODE process cannot be reversed perfectly in practice, the accumulated error is negligible, and DDIM inversion can nearly reconstruct the original image (see ``CFG ($\omega\!=\!0.0$)'' in Fig.~\ref{F3_1}). 
However, to generate an image well aligned with the conditional text prompt using Stable Diffusion, a large guidance scale $\omega\!>\!1$ is necessary for the sampling process in Eq.~\ref{E4}. This arouses the problem that when enlarging $\omega$, the generated images are far from the original ones as shown in Fig.~\ref{F3_1}.
We believe that when $\omega$ of the sampling process is different from that of the forward process, the accumulated error would be amplified, leading to unsatisfactory reconstruction quality. This can be illustrated in Table~\ref{T1}, \emph{i.e.}, the best reconstruction quality in each line is obtained when $\omega$ used in the DDIM sampling process is the same as that used in the forward process. Even if $\omega$ used in the forward and sampling processes are kept the same, PSNR still decreases with $\omega$ increasing (see the numbers in bold in Table~\ref{T1}). The above analysis demonstrates that it is hard for DDIM inversion to achieve a satisfactory trade-off between editability (which requires larger $\omega$) and fidelity (which requires smaller $\omega$). To address this issue, we propose a new inversion technique, \emph{i.e.}, Prompt Tuning Inversion.
% To generate an image well aligned with the text prompts using the Stable Diffusion model, a large guidance scale $\omega\!>\!1$ is necessary. However, when enlarging $\omega$, we find that the generated images are far from the original ones as shown in \textcolor{red}{Fig.~\ref{F3_1}.} 
% We believe that the different guidance scales in the forward and sampling process amplify the accumulated error. The results in \textcolor{red}{Table~\ref{T1} show that with $\omega$ increasing, PSNR decreases, which
% demonstrates our opinions.} 

%------------------------------------------------------------------------

\subsection{Prompt tuning for inversion}
\label{sec:3.3}
To successfully invert real images into the model's domain, recent works optimize the textual encoding, the network's parameters, or both. Motivated by Pivotal Inversion~\cite{roich2022pivotal}, we replace the conditional embedding of the text prompt with an optimized one, referred to as \textbf{Prompt Tuning} in this work. Namely, for each input image, we optimize only the conditional embedding $c$
so that it encodes important information of the input image which helps the reconstruction.
% , which is initialized with the unconditional embedding or the target embedding. 
The parameters of the diffusion network and the text encoder $\tau_{\theta}$ are frozen during prompt tuning.

We first initialize $z_{0}^{*}\!=\!z_0\!=\!ENC(x_0)$, and adopt DDIM inversion with $\omega\!=\!0$ to obtain a trajectory of noisy latent codes $\{z_t\}_{t=1}^T$.
% $z_{0}^{*}$, ..., $z_{T}^{*}$. 
Then we initialize $\Tilde{z}_T=z_T^*$ and perform the following optimization
to the conditional embedding $c$
with $\omega\!>\!1$ for the timestamps $t\!=\!T,...,1$, each for $N$ iterations:
\begin{equation}
    \label{E7}
    c_t = \underset{c_t}{\arg\min}~||z^*_{t-1}-z_{t-1}(\Tilde{z}_t, t,c_t)||^2.
\end{equation}
For brevity, $z_{t-1}(\Tilde{z}_t,t,c_t)$ denotes applying a DDIM sampling step using $\Tilde{z}_t$ and the conditional embedding $c_t$ at the timestep $t$. At the end of each timestep, we update
\begin{equation}
    \label{E8}
    \Tilde{z}_{t-1} = z_{t-1}(\Tilde{z}_t,t,c_t).
\end{equation}
Finally, we can reconstruct the input image by using the noise latent $\Tilde{z}_T=z_T^*$ and the optimized conditional embeddings $\{c_t\}^T_{t=1}$. In the next subsection, we will introduce the approach to editing images with the target text prompt and the conditional embeddings.

%------------------------------------------------------------------------
\subsection{Prompt tuning for editing}
\label{sec:3.4}
Since the sequence of
the
conditional embeddings $\{c_t\}^T_{t=1}$ is optimized to fully reconstruct the input image, we believe that these optimized conditional embeddings have contained the most information of the original image,
and thus ensure high fidelity.
To achieve the desired modification,
% we
these optimized embeddings are adopted
% use it 
to 
perform the editing
% apply the desired edit 
by advancing in the direction of the target text embedding $c^*\!=\!\tau_\theta(\mathcal{P}^*)$
to ensure good editability also.
More formally, in the second stage, we simply interpolate between the target embedding $c^{*}$ and the optimized $c_t$ linearly at each timestamp. For a given hyper-parameter $\eta\in(0, 1]$, we obtain
\begin{equation}
    \label{E8}
    c_t = \eta \cdot c^{*} + (1-\eta)\cdot c_t,
\end{equation}
where the first term ensures the effective editability corresponding to the semantic contents in the target text, while the second term guarantees a good reconstruction of the original image.
% which is the conditional embedding representing the desired edited image which balances editability and fidelity. 
The algorithm is presented in Lines 14-21 in Algorithm~\ref{A1}.
% where $\beta$ represents the learning rate.} 
Note that when $\eta\!=\!0$ or $\eta\!=\!1$, the output of our editing method is the reconstructed original image, or the output of the baseline DDIM-Edit, respectively.
% \textcolor{red}{When $\eta\!=\!0$ or $\eta\!=\!1$, the output of our editing method is the approximate original image or the output of the baseline DDIM-Edit, respectively.}

Intuitively,
% Basically, 
our editing method is to find an intermediate representation between the original image and the output of DDIM-Edit. For a desired modification, the intermediate representation is supposed to contain both the structural information of the source image and the semantic contents of the target text prompt.
Eq.~\ref{E8} is only one way to achieve this, which we refer to as \textbf{condition interpolation}.
% To achieve this, we choose to change the conditional embedding through Eq.~\ref{E8} (referred to as \textbf{condition interpolation}). 
We also test a different interpolation method (referred to as \textbf{latent interpolation}), where we linearly interpolate between the noisy latent $z_t$ and $z_t^*$ at each timestamp:
\begin{equation}
    \label{E9}
    z_t^{edt} = \eta \cdot z_t + (1-\eta)\cdot z_t^*,
\end{equation}
where $z_t^*$ is the noisy latent calculated by DDIM inversion via Eq.~\ref{E6}, and $z_t$ is the latent obtained in the vanilla DDIM sampling process conditioned on the target embedding.
Although this approach is more simple since the process of prompt tuning is no longer needed,
% Although the editing method becomes more simple because the process of prompt tuning is no longer needed, 
we observe that this interpolation method may lead to cluttered images. This is because
the interpolation of latent embeddings mixes the source object and the edited object spatially, rather than semantically (as condition interpolation does), leading to cluttered contents in images.
% the source object and edited object are mixed in the spatial space rather than the semantic space.  

\begin{algorithm}[t]
\LinesNumbered
\caption{Prompt Tuning Inversion for Editing}
\label{A1}
\SetAlgoLined
\KwIn{An input image $\mathcal{I}$ and a target prompt embedding $c^*=\tau_\theta(\mathcal{P}^*)$
% = \tau(P^*)$.
}
\KwOut{Edited image $\mathcal{I}^*$.}

\tcp{DDIM Inversion}
Set guidance scale $\omega = 0, z_{0}^{*} = ENC(\mathcal{I})$;\\
Compute the intermediate
trajectory $\{z_t^*\}_{t=0}^T$
% results $z_{T}^{*}$, ..., $z_{0}^{*}$ 
using DDIM inversion over $\mathcal{I}$ without conditional guidance via Eq.~\ref{E6};\\

\tcp{Prompt Tuning}
Set guidance scale $\omega > 1$, $\eta \in (0, 1]$; \\
Initialize $\Tilde{z}_{T} \leftarrow z_{T}^{*}$,  $c_T \leftarrow c^{*}$;\\
\For{t = T, T-1, ..., 1}{
    \For{j = 0, ..., N-1}{
        $c_{t} \leftarrow c_{t} - \beta\nabla_{c}||z_{t-1}^{*} - z_{t-1}(\Tilde{z}_{t}, t, c_t)||_{2}^{2}$; \\
    }
    $\Tilde{z}_{t-1} \leftarrow z_{t-1}(\Tilde{z}_{t}, t, c_t)$, $c_{t-1} \leftarrow c_t$
}

\tcp{Editing}
Set $z_T^{edt} \leftarrow z_{T}^{*}$; \\
\For{t = T, T-1, ..., 1}{
    $c_t = (1 - \eta)\cdot c_t + \eta \cdot c^*$; \\
    $z_{t-1}^{edt} = z_{t-1}(z_{t}^{edt}, t, c_t)$; \\
}
$\mathcal{I}^* = DEC(z_0^{edt})$; \\
return $\mathcal{I}^*$
\end{algorithm}

\subsection{Discussion}

Our proposed image editing method shares similar motivations with existing works~\cite{kawar2022imagic,valevski2022unitune,mokady2022null,hertz2022prompt},
all of which
% because they all
aim to modify an image in a text-driven and mask-free manner. However, our approach differs from them significantly in the following aspects:

\textbf{1)} Imagic~\cite{kawar2022imagic} and UniTune~\cite{valevski2022unitune}
% In Imagic~\cite{kawar2022imagic} and UniTue~\cite{valevski2022unitune}, they must 
finetune the diffusion models for hundreds of steps to maintain high fidelity to the input image. In contrast, we only need to optimize the conditional embedding,
which greatly reduces computational budgets.

\textbf{2)} Our proposed Prompt Tuning Inversion is inspired by the Null-Text Inversion method~\cite{mokady2022null}. However, Null-Text Inversion chooses to optimize the unconditional embedding while we optimize the conditional one. Moreover, the editing process of Null-Text Inversion is achieved by the cross-attention map control in Prompt-to-Prompt~\cite{hertz2022prompt}, which requires an additional description of the input image. Compared to theirs, our method only needs the target text prompt
and is thus more user-friendly.

%------------------------------------------------------------------------
\section{Experiments}
% \subsection{Implementation Details}
\subsection{Setup}
\textbf{Implementation details. }In our experiments, we
adopt the text-conditional Latent Diffusion Model~\cite{rombach2022high} (known as Stable Diffusion) with 890M parameters trained on LAION-5B~\cite{schuhmann2022laion}
% use the 890M parameter text-conditional latent diffusion model trained on LAION-5B, known as Stable Diffusion, 
at $512\times 512$ resolution. 
For the DDIM schedule, we adopt 50 steps and retain the original hyper-parameter choices of Stable Diffusion. The encoding ratio parameter is set to $0.8$. 
The number of iterations to optimize $c$ per diffusion step is set to $N\!=\!1$. The hyper-parameters $\beta$ and $\eta$ in Algorithm~\ref{A1} are set to $0.1$ and $0.9$, respectively, unless specified. These allow editing an image in $\sim\!1$ minute on a single Tesla V100 GPU.
For better performance, we adopt attention maps to localize the edited regions (referred to as local blending), and re-weight the attention maps as in~\cite{hertz2022prompt}.
% \textcolor{red}{For better results, we use attention maps to localize the edited regions (referred to as local blending) and re-weight the attention maps as~\cite{hertz2022prompt}.}
% The number of iterations per diffusion step, $N$ is set to 1 and the learning rate is set to $0.1$. These allow editing images in ~1 minute on a single Tesla V100 GPU. 

% \subsection{Evaluation and Datasets}
\textbf{Evaluation and datasets. }In semantic image editing,
the visual content of the edited image is supposed to align well with the target text prompt (editability) while staying close to the input image in terms of the unedited parts (fidelity).
% we have to satisfy two contradictory objectives, editability and fidelity. The former means the visual content of edited images is supposed to align with the target text prompt. The latter means that the edited images should stay close to the input image. 
For a given method, better editability usually comes at the cost of decreased fidelity to the input image, and vice versa.
This forms a trade-off curve between the two objectives.
% By controlling the editability, image editing methods get different operating points, forming a trade-off curve between the two objectives aforementioned.
Following DiffEdit~\cite{couairon2022diffedit}, we evaluate different editing methods by comparing their trade-off curves
on ImageNet~\cite{deng2009imagenet}. Specifically, given
% , and we perform quantitative experiments on ImageNet. Given 
an image
of
% belonging to 
one class from ImageNet, we aim to edit it to
% translate it to 
another class of ImageNet as
instructed
% indicated 
by the target text prompt.
The editability and fidelity are measured using the LPIPS perceptual distance~\cite{zhang2018unreasonable} and CSFID, which is a
class-conditional FID metric~\cite{heusel2017gans}, respectively.
% Considering the class translation task on ImageNet, we measure editability and fidelity with 
% the LPIPS perceptual distance and CSFID, respectively. 
The former measures the distance with the input image while the latter measures both image realism and consistency \emph{w.r.t.} the target class. For both metrics, lower values indicate better editing performance.
  
\subsection{Comparison with other methods on ImageNet}

\begin{figure}[tbp]
    \centering
    \includegraphics[width=0.8\linewidth]{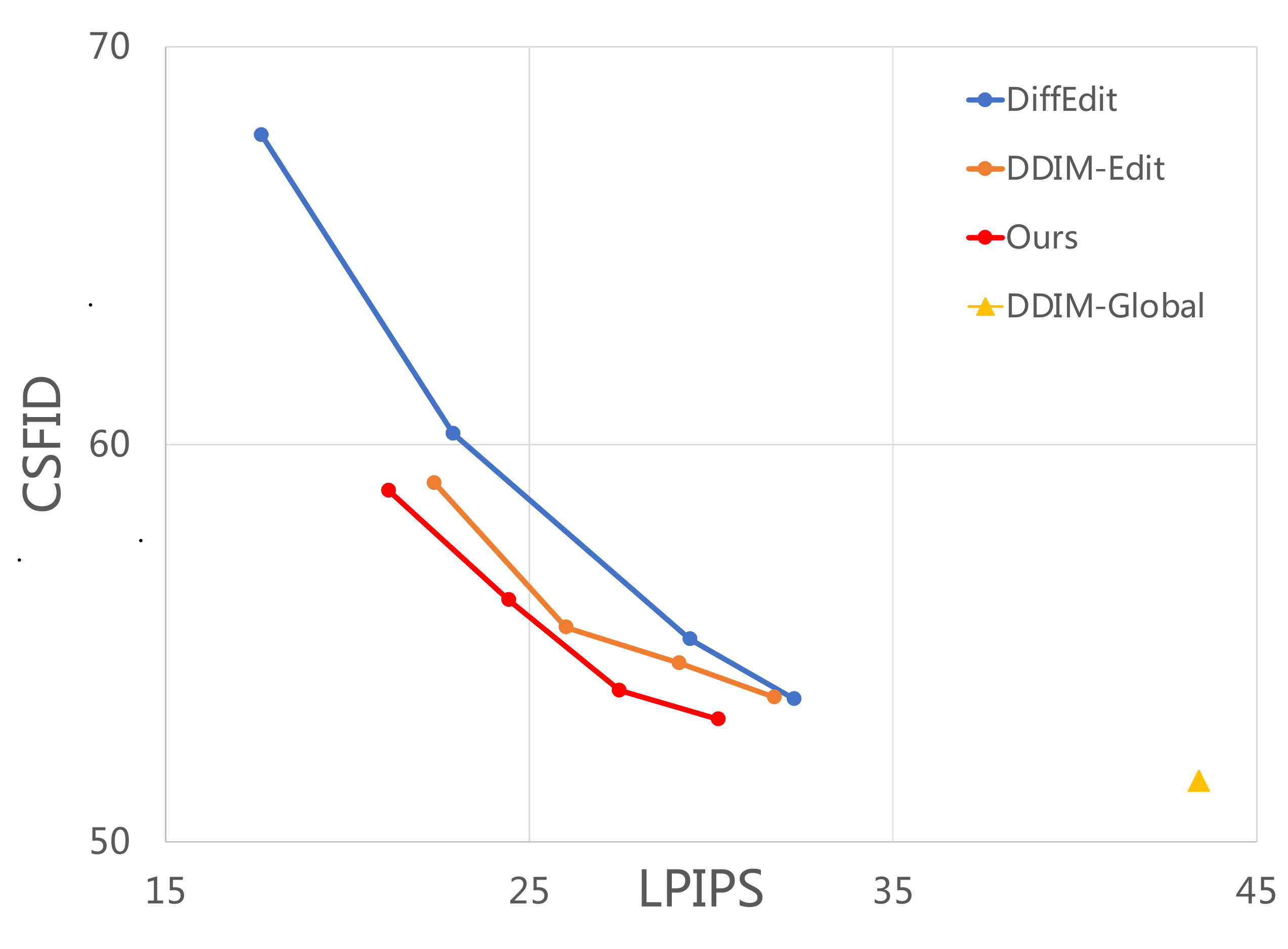}
    \caption{Comparison with DiffEdit and DDIM-Edit on ImageNet. For all methods, we set the DDIM encoding ratio to 0.8, and only vary the mask threshold to draw the trade-off curve.}
    \label{F4_1}
\end{figure}

We compare our method with DiffEdit and our baseline DDIM-Edit, since they both share the same DDIM forward process and a similar sampling process.
Besides,
they load
% these methods can use 
the same publicly available pre-trained weights for a fair comparison. 
To leverage the generalization capability of large-scale language-image models, we adopt the text-conditional Stable Diffusion model ``sd-v1-4'' instead of the class-conditional model trained on ImageNet as the pre-trained model.

\begin{figure}[t]
    \centering
    \includegraphics[width=0.95\linewidth]{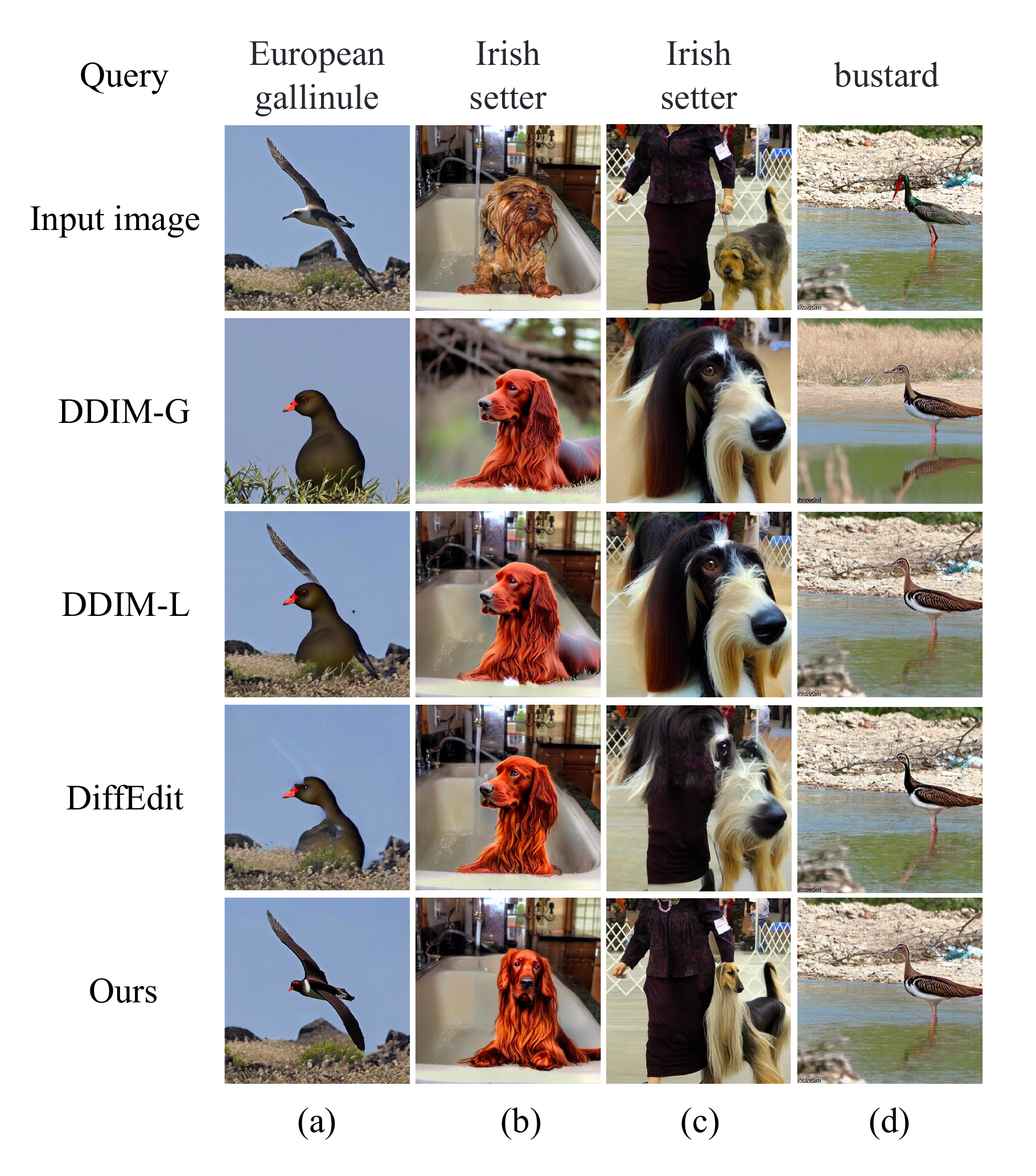}
    \vspace{-2mm}
    \caption{Editing examples on ImageNet by our method and other methods. DDIM-G/DDIM-L indicates the baseline method DDIM-Edit with/without the local blend trick.}
    \label{F4_2}
\end{figure}

As pointed out by DiffEdit~\cite{couairon2022diffedit}, different editing methods often have hyper-parameters which control editability, \emph{e.g.}, the mask threshold or the encoding ratio.
Lower mask threshold or higher encoding ratio leads to stronger editing.
% in ~\cite{couairon2022diffedit} and DDIM-Edit. 
In our proposed method,
% besides the above two parameters, 
we can also control the editing strength by varying the conditional interpolation ratio $\eta$ introduced in Eq.~\ref{E8}. 
In our evaluation, we fix the DDIM encoding ratio, and draw the trade-off curve by varying the mask threshold for all methods\footnote{As there was no official implementation of DiffEdit available at the time of writing, we adopted the unofficial implementation for inferring editing masks from https://github.com/LuChengTHU/dpm-solver.}.
% Therefore,  we fix the DDIM encoding ratio and use the mask threshold\footnote{As there is no official implementation available of DIFFEDIT at the time of writing, we used the unofficial implementation for inferring editing masks from https://github.com/LuChengTHU/dpm-solver.} to control the trade-off. 
The results are presented in Fig.~\ref{F4_1}, 
where  ``DDIM-Global'' denotes that the images are edited via ``DDIM-Edit'' but without using any masks, \emph{i.e.}, the editing is performed globally. This can be regarded as a lower bound of fidelity for all methods.
% where ``DDIM-Global'' denotes that edited images are generated without using any masks, which can be regarded as the upper bound of editability for all methods.
% The results are presented in Fig.~\ref{F4_1}. The \emph{DDIM-Global} means that edited images are sampled without using masks, which indicates the upper bound of the editability of all three methods. 
As the mask threshold increases, LPIPS decreases since fewer parts of the image are edited.
% With the mask thresholds increasing, we can observe that the LPIPS metric decreases due to more image modifications. 
Compared to DiffEdit, our baseline method DDIM-Edit achieves a better trade-off.
Note that the only difference between the two methods is the approach to generating masks. The comparison shows that inferring editing masks using cross-attention maps, as adopted by DDIM-Edit, is more appropriate.
Based on DDIM-Edit, our method can further improve the fidelity to the input images while maintaining editability. The best CSFID-LPIPS trade-off of our method demonstrates its superiority over DiffEdit and the baseline. 

We also present qualitative examples of these methods. As shown in Fig.~\ref{F4_2}, without automatically generated masks, ``DDIM-G'' tends to modify images globally. For simple cases (\emph{e.g.}, example (d)), image editing methods with the original DDIM inversion works well. However, for complex cases, we observe undesired and unreasonable edits to the objects. In contrast, with the help of the learnable conditional embedding, our method achieves realistic editing while successfully preserving the original details.

\subsection{Ablation study}
\noindent\textbf{Comparison to existing inversion methods.}
We randomly select 128 images and their corresponding captions from the COCO validation set~\cite{lin2014microsoft}.
% We randomly select 128 images and corresponding captions pairs from the COCO validation set. 
We then apply the following reconstruction methods to each image-caption pair: \textbf{(1) AE} denotes the variational auto-encoder with a slight KL-penalty used in Stable Diffusion.
An image is first encoded by the encoder of AE. Afterwards, the decoder directly reconstructs the image from the latent. Therefore, we consider AE as an upper bound of reconstruction quality. \textbf{(2) DDIM} denotes the DDIM inversion method, which is a baseline inversion method. As analyzed in Sec.~\ref{sec:3.2}, it usually outputs a low-quality reconstruction result under a large classifier-free guidance scale $\omega$. \textbf{(3) NTI} denotes the Null-Text Inversion method~\cite{mokady2022null}, which is our main point of comparison. Different from our method, it optimizes the unconditional embedding. \textbf{(4) PTI} denotes our proposed Prompt Tuning Inversion method. We introduce a learnable conditional embedding and the optimizing details are presented in Algorithm~\ref{A1}.

% \noindent\textbf{AE} denotes the variational auto-encoder with a slight KL-penalty used in Stable Diffusion. 
% An image is first encoded by the encoder of AE. Afterwards, the decoder directly reconstructs the image from the latent. Therefore, we consider AE as an upper bound of reconstruction quality.}
% In Stable Diffusion, an image is first encoded by the encoder of AE and the diffusion process is performed in a latent space. Afterwards, the decoder reconstructs the image from the latent. Therefore, we consider AE as an upper bound of reconstruction quality.

\begin{table}[t]
    \centering
    \small
    \resizebox{0.85\linewidth}{!}{
    \begin{tabular}{c|c|c|c|c}
         \hline
         Method & AE     & DDIM   & NTI    & \textbf{PTI (ours)} \\
         \hline
         PSNR   & 26.22  & 13.64  & 24.45  & 25.71  \\
         \hline
         SSIM   & 0.8564 & 0.4641 & 0.8270 & 0.8501 \\
         \hline
    \end{tabular}
    }
    \vspace{+1mm}
    \caption{Reconstruction quality of different methods measured by PSNR and SSIM. For both metrics, higher values indicate better quality.}
    \label{T2}
\end{table}

\begin{table}[t]
    \centering
    \small
    \resizebox{0.85\linewidth}{!}{
    \begin{tabular}{c|c|c|c|c|c}
         \hline
         iters & 1      & 2      & 3      & 4     & 5     \\
         \hline
         \multicolumn{6}{c}{learning rate $\beta=0.01$}         \\
         \hline         
         NTI   & 17.05  & 20.12  & 22.25  & 23.61 & 24.45 \\
         \hline
         PTI   & 19.36  & 23.20  & 24.86  & 25.47 & 25.71 \\
         \hline
         \multicolumn{6}{c}{learning rate $\beta=0.1$}          \\
         \hline         
         NTI   & 23.08  & 24.74  & 25.62  & 25.82 & 25.91 \\
         \hline
         PTI   & 24.74  & 25.23  & 25.78  & 25.90 & 25.97 \\
         \hline
    \end{tabular}
    }
    \vspace{+1mm}
    \caption{PSNR scores under different optimizing settings.}
    \label{T3}
\end{table}

% \noindent\textbf{DDIM} denotes the DDIM inversion method, which is a baseline inversion method. As analyzed in Sec.~\ref{sec:3.2}, it usually outputs a low-quality reconstruction result under a large classifier-free guidance scale $\omega$.

% \noindent\textbf{NTI} denotes the Null-Text Inversion method~\cite{mokady2022null}, which is our main point of comparison. Different from our method, it optimizes the unconditional embedding. 

% \noindent\textbf{PTI} denotes our proposed Prompt Tuning Inversion method. We introduce a learnable conditional embedding and the optimizing details are presented in Algorithm~\ref{A1}.

The experimental results are provided in Table~\ref{T2}.
For the diffusion-based inversion methods, we apply the diffusion model in an unconditional manner (\emph{i.e.}, the classifier guidance scale $\omega\!=\!0$) for the DDIM forward process. For the sampling process, we set $\omega\!=\!7.5$. 
As shown in Table~\ref{T2},
% due to the different $\omega$, 
DDIM inversion fails to reconstruct the original images since $\omega$ in the sampling process is different from that in the forward process, leading to a low PSNR score ($13.64$).
% , and its PSNR score is only $13.64$. 
For NTI and PTI, we set the number of iterations $N$ to 5 and the learning rate $\beta$ to 0.01. We observe that both methods can reconstruct the images but the reconstruction quality of our method is better (25.71 vs. 24.45). To further demonstrate the effectiveness of our method, we vary $N$ from $1$ to $5$ and increase the learning rate $\beta$ from 0.01 to 0.1. The experimental results in Table~\ref{T3}
shows that
% . We observe that 
the reconstruction quality of PTI is always better than NTI under all settings, demonstrating that our method converges faster.

\begin{figure}[t]
    \centering
    \includegraphics[width=0.485\textwidth]{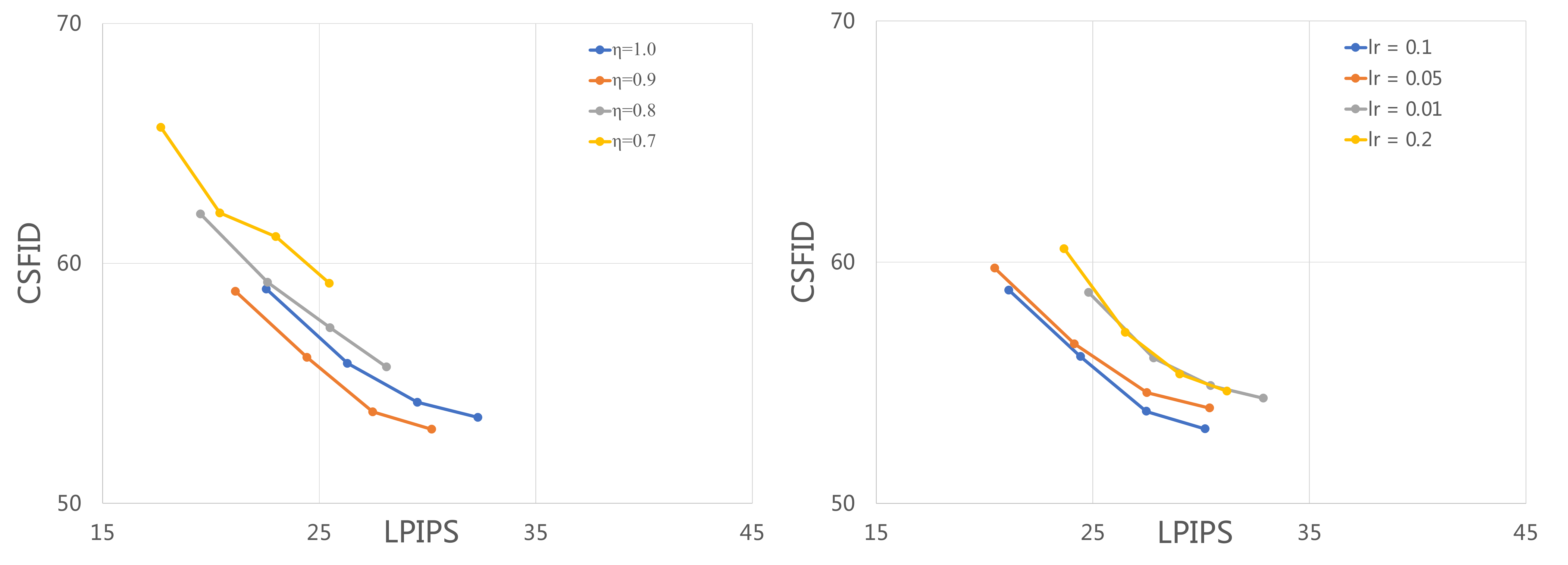}
    \caption{\textbf{Left}: ablation on the
    % influence of 
    interpolation ratio $\eta$. \textbf{Right}: ablation on the learning rate $lr$ (\emph{i.e.}, $\beta$) in Prompt Tuning Inversion.}
    \label{F4_3}
\end{figure}

\noindent\textbf{Influence of other hyper-parameters.}
We also
perform ablation on
% ablate 
two core components of our method, \emph{i.e.}, the interpolation ratio $\eta$ and the learning rate $\beta$ in PTI, to measure their influence in terms of CSFID-LPIPS on ImageNet. When $\eta\!=\!1$ or $\beta\!=\!0$, our method reverts to the baseline method DDIM-Edit. The left panel of Fig.~\ref{F4_3} shows decreasing $\eta$ from $1.0$ to $0.9$ leads to a better CSFID-LPIPS trade-off but lower ratios result in a worse balance between editability and fidelity. When we fix $\eta$ as $0.9$ and decrease $lr$ from $0.1$ to $0.05$ or $0.01$, the trade-off also becomes worse.  

%------------------------------------------------------------------------

\begin{figure}[t]
    \centering
    \includegraphics[width=0.485\textwidth]{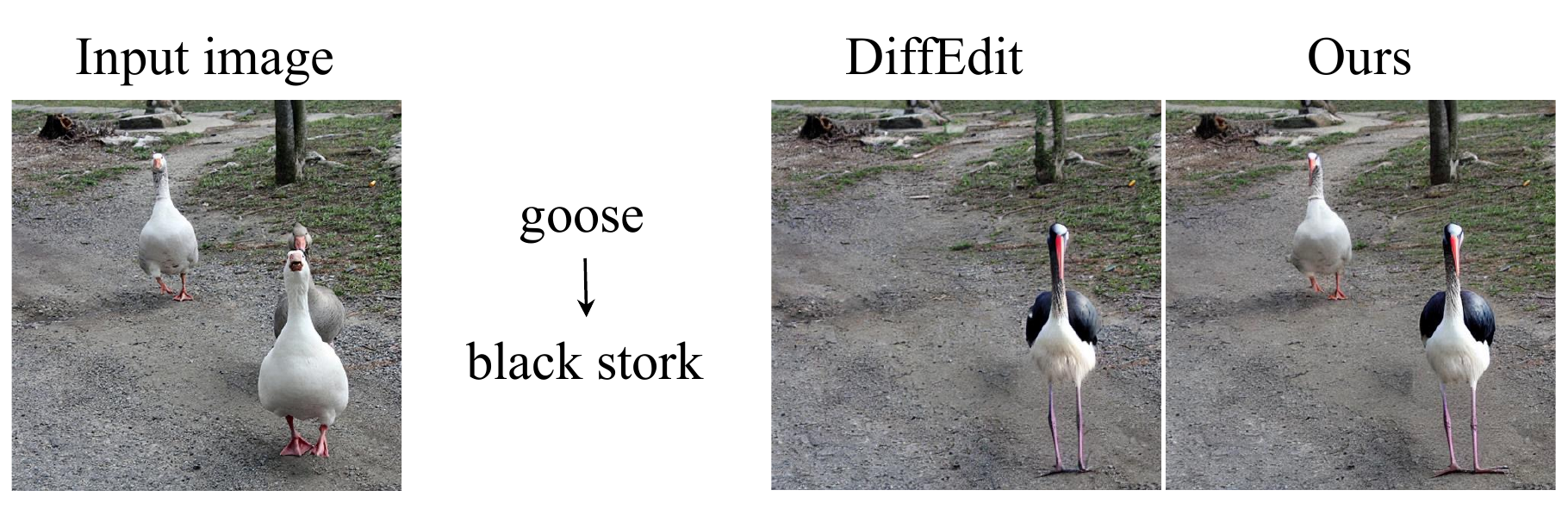}
    \caption{A failure example on ImageNet: when multiple objects exist, only one of them is editied successfully.}
    \label{F4_4}
\end{figure}

\section{Conclusion and future work}
We propose an intuitive and user-friendly text-based image editing method, which benefits from the superior generation and generalization capacities of large-scale image language diffusion models (\emph{e.g.}, Stable Diffusion). The key idea of our method is that
important structural information of the input image can be encoded into conditional embeddings, which can guide the diffusion model to reconstruct the original image via the sampling process. Based on this,
% images can be inverted into the corresponding conditional embeddings, which can guide the diffusion model to reconstruct the original images so that we can encode the structural information into these embeddings via a reconstruction process. 
% Thus 
our method consists of two stages. In the first reconstruction stage, we propose a novel Prompt Tuning Inversion method which encodes image information to learnable conditional embeddings quickly and accurately. In the second editing stage, we introduce an interpolation which linearly combines the target text embedding with the optimized embedding obtained in the first stage. In this way, the new conditional embedding contains both information from the input image and the target text prompt, resulting in an edited image with an appropriate trade-off between editability and fidelity. Quantitative and qualitative experimental results show that our approach achieves superior editing performance of images over previous methods. 

While our method works well in most scenarios, it still faces some limitations. As shown in Fig.~\ref{F4_4}, there are multiple objects in the input images. However, neither DiffEdit nor our method changes all ``gooses'' to ``black storks''. This limitation can possibly be mitigated by operating the attention maps more precisely~\cite{chefer2023attend} or adding different modes of conditional control~\cite{zhang2023adding}, providing a research direction for image editing. We leave these options for future work.

%------------------------------------------------------------------------
{\small
\bibliographystyle{ieee_fullname}
\bibliography{diffusion}
}

\end{document}